\documentclass[11pt,a4paper]{article}
\usepackage[hyperref]{acl2017}
\usepackage{times}
\usepackage{latexsym}
\usepackage{graphicx}
\usepackage{appendix}

\usepackage{url}

\aclfinalcopy % Uncomment this line for the final submission
\title{CMPUT 600 Project Report: Homonym Identification using BERT}

\author{Rohan Saha \\
  Department of Computing Science / University of Alberta \\
  {\tt rsaha@ualberta.ca}}

\begin{document}
\maketitle
\begin{abstract}
 Homonym identification is an important for WSD that require coarse grained partitions of senses. The goal of this project is to determine whether contextual information is sufficient for identifying a homonymous word. To capture the context, BERT embeddings are used as opposed to Word2Vec, which conflates senses into one vector. SemCor is leveraged to retrieve the embeddings. Various clustering algorithms are applied on the embeddings. Finally, the embeddings are visualized in a lower-dimensional space to understand the feasibility of the clustering process.
\end{abstract}

\section{Introduction}
The study of homonymy is often debated and may be considered to be undifferentiated to polysemy by many. Nevertheless, the study of homonymy is important for many tasks in lexical semantics\cite{Navigli2017}. Some tasks in lexical semantics depend on how fine-grained the lexical resources are. For example, in some cases, it may not be possible to use WordNet\cite{wordnet1998fellbaum} because many studies have shown that the distinction among senses in WordNet is too fine grained and work has also been done to merge senses that are similar resulting in a coarse grained sense inventory\cite{Snow2007}. \par The task of homonym identification can also help to indirectly alleviate this problem because the task requires the clustering of words types into different groups based on the semantic similarity of the word senses. An example of a homonym would be the word type 'bank' shown in table \ref{table:bank_homonym_example}\cite{Hauer2019}.

As a formal definition for the task, given a word $w$, classify $w$ as a homonymous or a non-homonymous word. Loosely, the task is formulated as a binary classification problem. For the purpose of explaining the methodology and the results, the word 'light' will be used for $w$. 'Light' has three homonymous senses - illumination, skinny, and lessen. The goal is to devise an experiment that can identify at least two of these three homonymous senses which will classify $w$ as a homonymous word. Similarly if the procedure when applied on $w$ results in only the identification of only one homonymous sense, then $w$ is not a homonym.\\ \\
The methodology described in section \ref{section:proposed} leverages contextual information represented in the form of token level embeddings. Using these contextual embeddings, the hope is to observe more than one cluster for a homonymous word in an $n$-dimensional space, where $n$ is the number of dimensions of the contextual embedding, and zero or no clusters for a non-homonymous word.
\begin{table}[h!]
    % \centering
    \begin{tabular}{|l|l|  c|  c| } \hline
    \multicolumn{1}{|c|} {Bank(n1)} & \multicolumn{1}{|c|} {Bank (n2)} \\ \hline
 \#2 financial institution & \#1  sloping land\\
 \#5  stock held in reserve & \#3 long ridge or pile\\
 \#6  funds held by a house & \#4 arrangement of objects\\
 \#8 container for money & \#7 slope in a road\\
 \#9 building & \#10  flight maneuver\\ \hline
    \end{tabular}
    \caption{The senses of the noun ‘bank’ from WordNet 3.0,
grouped by its two homonyms with respective sense numbers. Bank(n1) represents the first homonym of 'Bank' that is a noun. Bank(n2) represents the second homonym of 'Bank' that is a noun.}
    \label{table:bank_homonym_example}
\end{table}

\section{Related Work}
\label{section:task}
Most of the previous work for homonym detection uses embeddings that are not truly contextual, for example, embeddings from Word2Vec. The most closely related work on homonym detection is the work of \cite{van-den-beukel-aroyo-2018-homonym} where humour in short text is recognized by integrating homophone and homograph based features into existing humour recognition systems. Another work tries to the understand the semantic classification of homonyms with respect to human understanding by finding the most suitable computational model. One other interesting work address the importance of distinguishing polysemy and homonymy in information retrieval using morphology, part-of-speech, and phrases\cite{krovetz-1997-homonymy}. Though this project address clustering which is an unsupervised machine learning technique, the work of \cite{Wiedemann2019} uses contextual embeddings from BERT, ELMo and Flair NLP to classify WordNet senses using a supervised classification algorithm (K-Nearest Neighbours).

\section{Background Knowledge}
\label{section:background}
The experiment uses various clustering and dimensionality reduction techniques on the embeddings. In addition, the state-of-the-art BERT model is also used to obtain the contextual embeddings.
\subsection{BERT}
BERT\cite{Devlin2018} is a transformer based model that uses an encoder to map the input data (words) to a numerical vector or an embedding. It uses an attention mechanism to account for all the surrounding words of the target word. Therefore, one word token may have a different embedding depending on its intended meaning in the sentence. BERT embeddings are appropriate for the task given that it deals with polysemes. Sense disambiguation for a polyseme has been previously been carried out using BERT\cite{Du2019} and here BERT is used to distinguish between word types in a more coarse-grained manner than polysemes. In this project, a simplified API\footnote{\url{https://github.com/imgarylai/bert-embedding}} for BERT is used where the input is a sentence, and the output is the contextual embeddings for each of the word tokens in that sentence.
\subsection{Clustering Techniques}
Three clustering methods were used on the contextual embeddings. First was a hierarchical clustering known as Agglomerative Clustering\footnote{\url{https://scikit-learn.org/stable/modules/generated/sklearn.cluster.AgglomerativeClustering.html}}. Agglomerative Clustering mergers pairs of observations with the goal of minimally increasing a given linkage distance. In the experiment, the variance of the clusters being merged was minimized.\par The second clustering method was the MeanShift algorithm\footnote{\url{https://scikit-learn.org/stable/modules/generated/sklearn.cluster.MeanShift.html}} which works by updating candidates for centroids to be the mean of the clusters. The third algorithm that was used is the DBScan algorithm\footnote{\url{https://scikit-learn.org/stable/modules/generated/sklearn.cluster.DBSCAN.html}}. DBScan is a density based clustering algorithm that clusters observations with high spatial density in n-dimensions.
\subsection{Dimensionality Reduction Techniques}
In addition to the clustering methods, various dimensionality reductions techniques were used. Techniques used in this project were T-SNE, PCA, Isomap, locally linear embedding, and multi-dimensional scaling\footnote{\url{https://scikit-learn.org/stable/modules/manifold.html}}. Among these techniques, only the results using T-SNE are shown in the paper as other methods produced visualizations using which no significant inferences could be derived.

\section{Experiment Design and Approach}
\label{section:proposed}
The approach to addressing the task involves a series of steps starting from pre-processing the dataset to clustering the contextual embeddings.
\subsection{Pre-processing and Homonym mapping}
First, a cleaned up version of SemCor is used, which is a subset of the full data set used in \cite{Raganato}. First, all the sentences from SemCor are filtered out that do not match a Lemma in the list of resource homonyms file\footnote{\url{https://webdocs.cs.ualberta.ca/~kondrak/homonyms.html}} All the remaining target words in the cleaned up version of SemCor were then mapped to their corresponding sense key in WordNet. This mapping is required for an averaging process of the embeddings that is be explained later. Now, for each of the target words in this partially processed dataset, the homonymous groups need to be mapped. In other words, the polysemous senses of the target word $w$ in each sentence in SemCor need to be assigned a homonymous group. To do this, a resource mapping file containing the homonym group for each sense number is used$^6$. This file contains the WordNet sense numbers for each word type and the corresponding group. First, the partially processed dataset is mapped to the sense numbers for target words and then using the sense numbers, the homonymous group are assigned. Some homonymous senses are extremely rare and a corresponding sentence may not appear in SemCor. Such homonyms are not considered because the clustering algorithm will have no basis do distinguish between groups of homonyms. Therefore, sentences that are assigned only one group of homonyms for a target word are filtered out from SemCor.\par The final list of homonyms consisted only 37 homonymous word types for which SemCor contained sentences that represent at least two or more groups of homonyms.
\subsection{BERT Embeddings}
Using the pre-processed dataset, BERT embeddings are retrieved for the sentences in SemCor. The input supplied to the pre-trained BERT contains only 10 words on either side of the target word in the sentence because the API is not able to process sentences of greater length. Ten words usually preserves contextual information and there is no information loss. All the features of the embeddings are used for clustering because it is fundamental to minimize any information loss.\par The obtained embeddings are not directly fed into the clustering algorithm, instead they go through a averaging process. For each target word $w$, the embeddings are grouped according to the sense key. For example for sense key $a$, there may be five sentences where the target word $w$ is mapped to sense key $a$. All the contextual embeddings for these five instances of $w$ are averaged to create one contextual embedding. Therefore, for
each word $w$ the number of averaged embeddings is equal to its number of senses in SemCor. As a result, the number of embeddings is equal to the number of senses for $w$.
\subsection{Clustering Contextual Embeddings}
Now, the averaged embeddings are used as input to the clustering algorithm. The algorithms used had one common feature in that the number of the clusters in undefined. Therefore, the clustering algorithm does not know how many clusters are present in the embeddings; the algorithm is dependent only on its parameters to group the embeddings. The clustering in general, does not segregate the embeddings into distinct groups. The assigned labels from the clustering algorithms were then compared to the true labels to observe clustering performance.
% 
% By uncommenting {\small\verb|\aclfinalcopy|} at the top of this
%  document, it will compile to produce an example of the camera-ready formatting; by leaving it commented out, the document will be anonymized for initial submission.  When you first create your submission on softconf, please fill in your submitted paper ID where {\small\verb|***|} appears in the {\small\verb|\def\aclpaperid{***}|} definition at the top.

% The review process is double-blind, so do not include any author information (names, addresses) when submitting a paper for review.  
% However, you should maintain space for names and addresses so that they will fit in the final (accepted) version.  The ACL 2017 \LaTeX\ style will create a titlebox space of 2.5in for you when {\small\verb|\aclfinalcopy|} is commented out.  

\section{Results and Analysis}
\label{results}
The results are shown in this section for the word $w=light$. There was no specific motive to choose 'light' but only the fact that there were sentences in SemCor that represented at least two homonymous senses namely $illumination$ and $skinny$ corresponding to groups $100$ and $200$ in the resource homonyms$^6$. \par Here, only the results for Hierarchical Clustering is shown because among the three clustering algorithm used, Hierarchical Clustering actually assigns clusters to the points, whereas other algorithms mostly treated all points as outliers. Hierarchial Clustering tries to create clusters equal to number of actual groups of homonyms($hom\_type$). All the results can be replicated using the corresponding code\footnote{\url{https://github.com/simpleParadox/CMPUT-600-Project}}. Different dimensionality reduction techniques are also used here that help to analyze the visualization.
\begin{figure}[h!]
    \centering
    \includegraphics[height=150px]{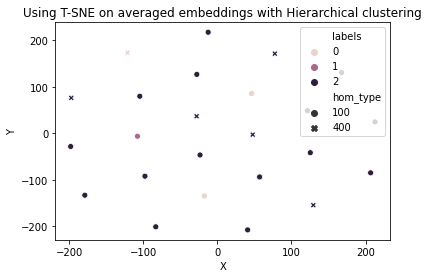}
    \caption{Results for Hierarchical Clustering with T-SNE}
    \label{fig:tsne_hierarchical}
\end{figure}
Figure \ref{fig:tsne_hierarchical} shows the results when using Hierarchical Clustering using T-SNE as a dimensionality reduction technique. It can be observed that when using T-SNE, the points are really spaced out. This is because T-SNE by default tries to compensate for 'crowding' of points when projecting to a lower dimensional space. It may be fair to argue that for the purposes of this task, T-SNE may not be a good method for visualization. Also, we can observe that the clustering algorithm classifies all the points into three different clusters where cluster label 1 has only one point. The cluster labels are the outcome of the clustering algorithm and $hom\_type$ is the actual true label(homonymous group) of the homonym. The $hom\_type$ are either homonymous groups 100 or 400 (for $w=light$), represented by solid circle and cross respectively. The labels are colored which show the cluster number from the clustering algorithm. Therefore, if the points are circle and black, then the true homonymous group is 100 and the cluster label is 2. If the point is black and a cross, then the true homonymous group is 400 but the clustering algorithm assigned it the label 2, and hence incorrect.

\begin{figure}[h!]
    \centering
    \includegraphics[height=150px]{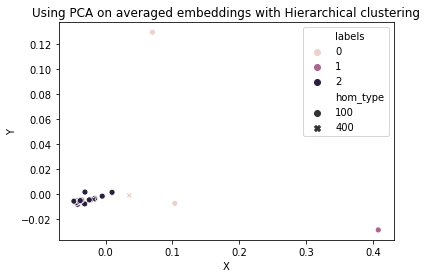}
    \caption{Results for Hierarchical Clustering with PCA}
    \label{fig:pca_hierarchical}
\end{figure}
In figure \ref{fig:pca_hierarchical}, the results for Hierarchical Clustering are shown with principal components analysis as a dimensionality reduction technique. In this case, the points seem to be close to each other in the lower dimensional space. PCA works by preserving only those variables that are independent of other variables and contribute to the highest variation in the data. From the visualization, it can be seen that there is a 'crowding' of points in one area of the figure and this is because PCA is a linear transformation, and does not account for any non-linearity in data. The figure shows that embeddings of homonym types 100 and 400 overlap and are are difficult to assign distinct clusters. And this is the reason for a 'crowding' of points at the bottom left of the figure; embeddings in group 100 and 400 are assigned cluster label 2 even though they should be assigned different cluster labels.

\begin{figure}[h!]
    \centering
    \includegraphics[height=150px]{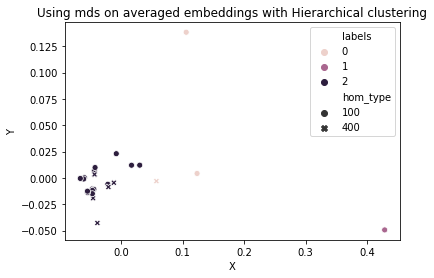}
    \caption{Results for Hierarchical Clustering with Multidimensional Scaling}
    \label{fig:mds_hierarchical}
\end{figure}
Figure \ref{fig:mds_hierarchical} shows the results for Hierarchical Clustering when visualized using a dimensionality reduction technique known as Multidimensional Scaling(MDS). Figure \ref{fig:mds_hierarchical} show groups of points in small area and thus the probable success of the clustering process but on close observation it can be seen the there are numerous points belonging to the homonymous group $100$ and $400$ being assigned cluster label 2. Therefore, figure \ref{fig:mds_hierarchical} may indicate the presence of clusters, but the clustering algorithm fails to assign correct labels. It must be noted that MDS works better than other dimensionality reduction techniques because it attempts to model similarity or dissimilarity of data as distances in higher dimensional spaces. Thus, the visualizations here are more indicative of similar embeddings than in figure \ref{fig:tsne_hierarchical}. However, it does not explain the inability of the clustering algorithm to assign correct label.\par From the visualizations and the analysis, it is easy to observe that the accuracy is below chance which raises doubts on the usage of a clustering approach for the task. In addition, from the experiment it was also observed that if the clustering algorithms were successful in assigning correct labels for one word using one hyperparameter setting, the same hyperparameters were unsuccessful for other words. Therefore, the algorithms lack generalization.

\section{Conclusion}
In this project, the problem of homonym identification was investigated where an experiment was performed to check the validity of a hypothesis underlying homonyms, that is contextual information in the form of word embeddings is sufficient for identifying homonyms. \par
From the experiment, it seems that the stated hypothesis may not be true owing to the inability of the clustering algorithms to assign distinct clusters in high-dimensional space. An explanation for such a finding is that though the senses in a homonymous group are semantically related (and are polysemous), the polysemes when represented in an n-dimensional space are almost evenly spaced out, and therefore provides a challenge for clustering the embeddings.\par An extension to the task will be to use a supervised learning algorithm using embeddings as features for classifying candidate word types into homonyms and non-homonyms. Though such a technique will require intensive effort in terms of data collection and preparation, a supervised learning algorithm may help create new baselines for any future research with regard to homonym identification.
\section{Acknowledgement}
I would like to thank Professor Grzegorz Kondrak for clarifying any questions and concerns about the project and organizing project meetings, which helped to monitor progress and address challenges. I would like to thank Bradley Hauer for compiling the list of known homonyms and also providing the cleaned up version of SemCor. In addition, my classmate Garnet Liam Peet-Pare was extremely helpful for the project with regard to mathematical aspects of the task.

%\bibliography{acl2017}
\bibliography{acl2017}
\bibliographystyle{acl_natbib}

\appendix
\section{Appendices}
\label{sec:appendix}
In the report, only the results from the Hierarchical Clustering method are shown various dimensionality reduction techniques for $w=light$. In the appendix, the visualizations from the remaining clustering techniques with various dimensionality reduction techniques.

\begin{figure}
    \centering
    \includegraphics[height=150px]{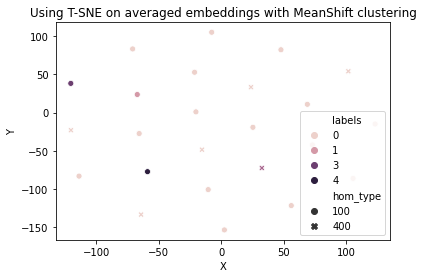}
    \includegraphics[height=150px]{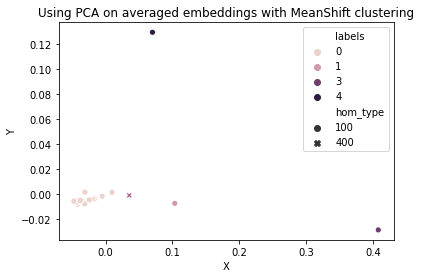}
    \includegraphics[height=150px]{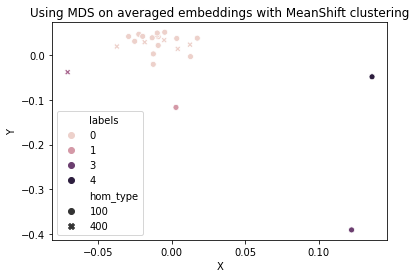}
    \caption{Results for MeanShift Clustering with T-SNE, PCA, and MDS for w=light}
    \label{fig:meanshift}
\end{figure}
Subfigure 1 in figure \ref{fig:meanshift} shows the visualization for MeanShift Clustering technique with T-SNE as a dimensionality reduction technique for the word $w=light$. It must be noted here that MeanShift assigns four clusters whereas Hierarchical clustering assigns three. It can also be observed from the figure that clustering fails to assign consistent labels with respect to the homonym types 100 and 400. Also, the points are really spaced apart because T-SNE respects the distances in higher dimensional space. \par Subfigure 2 in figure \ref{fig:meanshift} shows the visualization for MeanShift Clustering Technique for $w=light$ with PCA. Again initial observation may indicate presence of clusters, it can be seen that the labels are incorrect. Points belonging to different homonymous groups (denoted by cross and solid circle) are assigned the same labels and proves the inability of algorithm to find clusters.\par
\begin{figure}[h!]
    \centering
    \includegraphics[height=150px]{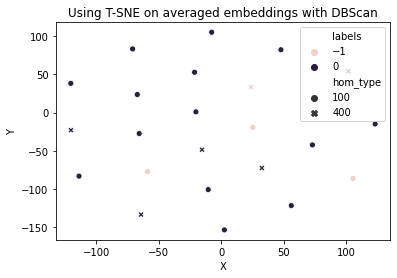}
    \includegraphics[height=150px]{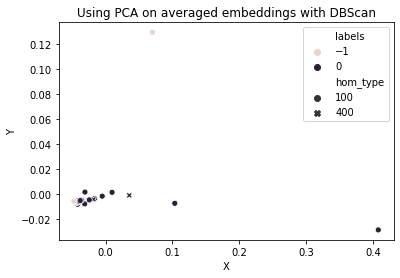}
    \includegraphics[height=150px]{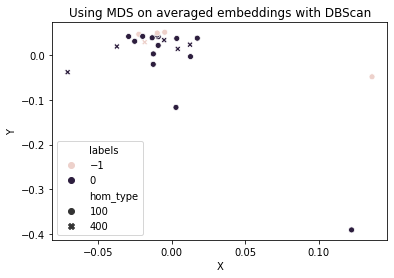}
    \caption{Results for DBScan with T-SNE, PCA, MDS for w=light}
    \label{fig:dbscan}
\end{figure}

 Subfigure 3 in \ref{fig:meanshift} shows the results for MeanShift Clustering with MDS. Again, similar to the results for PCA, the points are assigned the same cluster label even though they belong to distinct homonymous groups.
\\ \\
Figure \ref{fig:dbscan} shows results for $w=light$ with DBScan as the clustering algorithm.
For the DBScan algorithm, the visualizations are not significantly different than those for Hierarchical Clustering and MeanShift Clustering but only supports the inability of the algorithm to form clusters. \par
\begin{figure}
    \centering
    \includegraphics[height=150px]{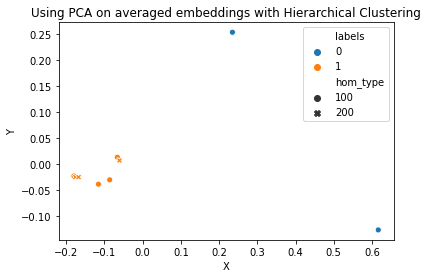}
    \includegraphics[height=150px]{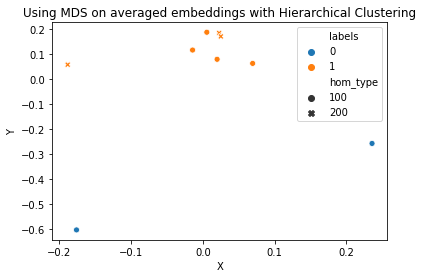}
    \caption{Results for Hierarchial Clustering with PCA and MDS for w=mean.}
    \label{fig:hierarchical_mean_pca_mds}
\end{figure}

\begin{figure}
    \centering
    \includegraphics[height=150px]{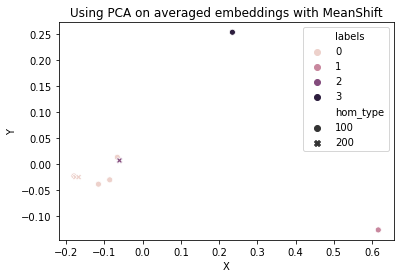}
    \includegraphics[height=150px]{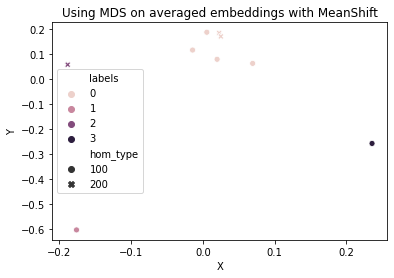}
    \caption{Results for MeanShift Clustering with PCA and MDS for w=mean.}
    \label{fig:meanshift_mean_pca_mds}
\end{figure}

\begin{figure}
    \centering
    \includegraphics[height=150px]{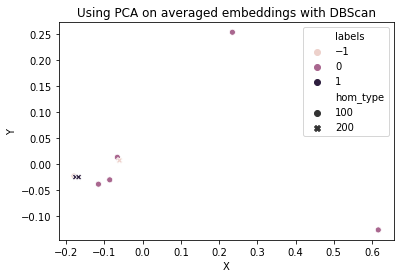}
    \includegraphics[height=150px]{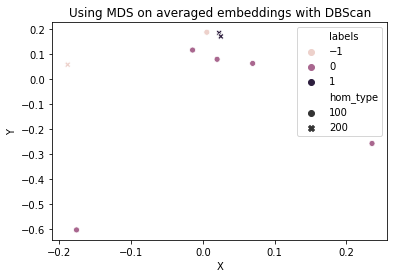}
    \caption{Results for DBScan Clustering with PCA and MDS for w=mean.}
    \label{fig:dbscan_mean_pca_mds}
\end{figure}

 Figure \ref{fig:hierarchical_mean_pca_mds}, figure \ref{fig:meanshift_mean_pca_mds}, and \ref{fig:dbscan_mean_pca_mds} for the word $w=mean$. 'mean' was chosen from the list of homonyms because SemCor contains sentences that represent at least two of the homonymous groups that is 100 and 200 for the homonymous senses $signify$ and $unkind$ respectively. It is useful to note that the number of sentences in SemCor for $w=mean$ is more than $w=light$. However, the actual number of senses is less compared to those of $w=light$(many sentences have the same sense for mean). As a result, the number of averaged embeddings is much lesser.\par Figure shows \ref{fig:hierarchical_mean_pca_mds} the results for Hierarchical clustering with PCA and MDS. We see that the points appear to be closer in a low dimensional space but the cluster labels are assigned incorrectly by the clustering algorithm. In figure \ref{fig:meanshift_mean_pca_mds} shows the results for the MeanShift clustering algorithm with PCA and MDS for $w=mean$. MeanShift performs worse than Heirarchical clustering because the number of cluster labels decided by the algorithm is more than the actual number of homonymous groups.
Figure \ref{fig:dbscan_mean_pca_mds} shows a similar visualization but the algorithm treats many points as outliers and thus assigns the label -1. \par All in all, Hierarchical Clustering (Agglomerative Clustering) performs relatively better than all the other clustering methods.

\end{document}